\documentclass{ieeeaccess}
\usepackage{cite}
\usepackage{amsmath,amssymb,amsfonts}
\usepackage{algorithmic}
\usepackage{graphicx}
\usepackage{textcomp}
\usepackage{url}
\usepackage{siunitx}
\usepackage[utf8]{inputenc}
\usepackage{nicefrac}

\usepackage[nolist]{acronym}

\usepackage{soul}

\begin{document}
	
\history{Date of publication xxxx 00, 0000, date of current version March 16, 2020.}
\doi{xx.xxxx/ACCESS.xxxx.DOI}

\title{MAT-Fly: an educational platform for simulating Unmanned Aerial Vehicles aimed to detect and track moving objects}

\author{\uppercase{Giuseppe Silano}\authorrefmark{1,2}, \IEEEmembership{Student Member, IEEE}, and \uppercase{Luigi Iannelli}\authorrefmark{2}, \IEEEmembership{Senior Member, IEEE}}

\address[1]{Faculty of Electrical Engineering, Czech Technical University in Prague; 16636 Prague 6 (Czech Republic) (e-mail: \texttt{giuseppe.silano@fel.cvut.cz})}

\address[2]{Department of Engineering, University of Sannio in Benevento, Piazza Roma 21; 82100 Benevento (Italy) (e-mail: \texttt{\{giuseppe.silano, luigi.iannelli\}@unisannio.it})}

\tfootnote{This work was partially funded by the ECSEL Joint Undertaking research and innovation programme COMP4DRONES under grant agreement no. 826610 and by the European Union's Horizon 2020 research and innovation programme AERIAL-CORE under grant agreement no. 871479.}

\markboth
{Silano \headeretal: MAT-Fly: an educational platform for simulating Unmanned Aerial Vehicles aimed to detect and track moving objects}
{Silano \headeretal: MAT-Fly: an educational platform for simulating Unmanned Aerial Vehicles aimed to detect and track moving objects}

\corresp{Corresponding author: Giuseppe Silano (e-mail: \texttt{giuseppe.silano@fel.cvut.cz}).}


\begin{acronym}
    \acro{BTH}[BTH]{Balanced Histogram Thresholding}
    \acro{CAMShift}[CAMShift]{Continuously Adaptive Mean-Shift}
	\acro{CVS}[CVS]{Computer Vision System}
	\acro{DOF}[DoF]{Degrees of Freedom}
	\acro{ESP}[ESP]{Electronic Stability Program}
	\acro{IB}[IB]{Integral Backstepping}
	\acro{IBVS}[IBVS]{Image-Based Visual Servoing}
	\acro{PID}[PID]{Proportional-Integral-Derivative}
	\acro{ROI}[ROI]{Region of Interest}
	\acro{SIL}[SIL]{Software-in-the-loop}
	\acro{UAV}[UAV]{Unmanned Aerial Vehicle}
	\acro{VR}[VR]{Virtual Reality}
\end{acronym}


\begin{abstract}
	
The main motivation of this work is to propose a simulation approach for a specific task within the~\ac{UAV} field, i.e., the visual detection and tracking of arbitrary moving objects. In particular, it is described MAT-Fly, a numerical simulation platform for multi-rotor aircraft characterized by the ease of use and control development. The platform is based on Matlab\textsuperscript{\textregistered} and the MathWorks\textsuperscript{\texttrademark}~\ac{VR} and~\ac{CVS} toolboxes that work together to simulate the behavior of a quad-rotor while tracking a car that moves along a nontrivial path. The~\ac{VR} toolbox has been chosen due to the familiarity that students have with Matlab and because it does not require a notable effort by the user for the learning and development phase thanks to its simple structure. The overall architecture is quite modular so that each block can be easily replaced with others simplifying the code reuse and the platform customization.

Some simple testbeds are presented to show the validity of the approach and how the platform works. The simulator is released as open-source, making it possible to go through any part of the system, and available for educational purposes.

\end{abstract}


\titlepgskip=-15pt

\maketitle

\section{Introduction}

\PARstart{U}{nmanned} Aerial Vehicles (UAVs), although originally designed and developed for defense and military purposes (e.g., aerial attacks or military air covering), in the recent years gained an increasing interest and attention related to civilian use. Nowadays,~\acp{UAV} are employed for several tasks and services like surveying and mapping~\cite{Scaramuzza2014}, for spatial information acquisition and buildings inspection~\cite{Choi2015}, data collection from inaccessible areas~\cite{Fraundorfer2012}, agricultural crops and monitoring~\cite{Anthony2014}, manipulation and transportation or navigation purposes~\cite{Blosch2010}. 

Many existing algorithms for the autonomous control~\cite{Castro2016} and navigation~\cite{Mancini2017} are provided in the literature, but it is particularly difficult to make the~\acp{UAV} able to work autonomously in constrained and cluttered environments or also indoors. Thus, it follows the need for tools that allow to understand what it happens when some new 
applications are going to be developed in unknown or critical situations. Simulation is one of such helpful tools, widely used in robotics~\cite{Abhijeet2015, Rosen2008, Silano2020SprinerBook, Sinha2021SMPT, Silano2019SMC}, whose main benefits are costs and time savings, enabling not only to create various scenarios, but also to carry out and to study complex missions that might be time consuming and risky in real world applications. Moreover, bugs and mistakes cost virtually nothing: it is possible to crash a vehicle several times and thereby getting a better understanding of implemented methods under various conditions. Thus, simulation environments are very important for fast prototyping and educational purposes, although they may have some drawbacks and limitations, such as the lack of noisy real data or the fact that simulated models are usually incomplete or inaccurate. Despite the limitations, the advantages that the simulation provides are more, as like as to manage the complexity and heterogeneity of the hardware, to promote the integration of new technologies, to simplify the software design, to hide the complexity of low-level communication~\cite{Elkady2012}.

Different solutions, typically based on external robotic simulators such as Gazebo~\cite{Koenig2004}, V-REP~\cite{VREP2013}, AirSim~\cite{airsim2017fsr}, MORSE~\cite{Echeverria2011}, are available. They employ recent advances in computation and computer graphics (e.g., AirSim is a photorealistic environment~\cite{Mancini2017}) in order to simulate physical phenomena (e.g., gravity, magnetism, atmospheric conditions) and perception (e.g., providing sensor models) in such a way that the environment realistically reflects the actual world. In some cases, those solutions do not have enough features that could allow to create large scale complex environments close to reality. On the other hand, when the tools provide such possibilities, they are difficult to use or they require high computational capabilities~\cite{airsim2017fsr}. Definitely, it comes out that simulating the real world is a nontrivial task, not only due to multiple phenomena that need to be modeled, but also because their complex interactions ask the user a notable effort for the learning and development phase. For all such reasons, having a complete software platform that makes possible to test different algorithms and control strategies for~\acp{UAV} moving in a simulated 3D environment is increasingly important both for the whole design process and for educational purposes. 

In this paper, it is presented a software platform in which detection, tracking and control algorithms can be evaluated and tested all together in a 3D graphical tool. Due to the simple implementation and the limited possibilities of interfacing it with dedicated middlewares (e.g., ROS~\cite{Quigley2009}, YARP~\cite{Metta2006}, GenoM~\cite{Mallet2002}), the proposed platform should be meant with an educational purpose. However, that does not imply a loss of generality nor makes the platform less important. Indeed, as highlighted in~\cite{Khan2017}, the use of interactive learning approaches allows students to improve their technical knowledge and communication skills, giving them the experience of what they will encounter in a real world environment. Therefore, the platform can be appreciated for its potentialities thanks to the advantages coming from the use of a~\ac{SIL} approach~\cite{Castro2016, Day2015, Shokry2009}. In other words, the functionalities provided by the simulator can be easily expanded by students, researchers, and developers modifying or integrating new vehicles dynamics (e.g., hexarotor~\cite{Unicomb2017IROS}, fully actuated platform~\cite{Ryll2017ICRA}), control algorithms (e.g., geometric control laws~\cite{Lee2010CDC}, flatness-based control methods~\cite{Faessler2018}) or detection and tracking techniques (e.g., YOLO~\cite{Redmon2016CVPR, Jiao2019Access}, Fast R-CNN~\cite{Girshick2015ICCV, Jiao2019Access}) for their purposes.

Compared to the commercial and open-source platforms available in the literature~\cite{Baca2020mrs, Pignaton2020ICUAS, SanchezLopezJINT2017, LimRAM2012, Furrer2016}, the proposed framework runs on a built-in environment (i.e., Matlab and its toolboxes) and has no constraints in terms of hardware (e.g., memory, unit processor, etc.). Moreover, the simulator is self-contained (i.e., everything is in one place) and can also be used by people without programming skills (i.e., algorithms are typically written in the most common programming languages). Matlab and the~\acf{CVS}\footnote{\url{https://www.mathworks.com/products/computer-vision.html}} and~\acf{VR}\footnote{\url{https://www.mathworks.com/products/3d-animation.html}} toolboxes are the only tools the user needs to work with.

The specific domain of interest regards the behavior of multi-rotor aircraft acting in accordance with the~\ac{IBVS} approach~\cite{Chaumette2006, Chaumette2007}. The \textit{eye-in-hand} camera configuration~\cite{Siciliano2009} along with the pinhole camera model is considered for the aerial vehicle. Compared to other approaches~\cite{Lippiello2005}, the camera is rigidly attached to the~\ac{UAV} frame and moves according to the aircraft motion.
\Figure[!t]()[width=0.48\textwidth]{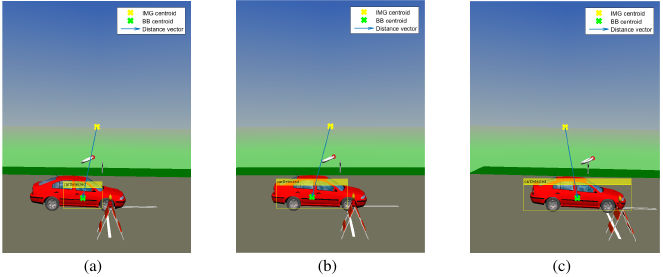}{Three consecutive frames produces as output by the~\acs{CAMShift} algorithm while tracking the target. The image and the bounding box centroids as well as the distance vector among centroids are depicted (see Sec.~\ref{sec:visionBasedTargetDetection}).\label{fig:camshiftTrackingAlgorithm}}

The application that is considered is an extension of the authors' previous work~\cite{Silano2016}, that has been revised for making the aircraft able to detect and track a specific object (a car) moving along a nontrivial path. This simple scenario is used as a testbed to show: (i) how the platform works, (ii) the elements that make up the software architecture, and (iii) the adaptability of the platform to the different needs of the user. Compared to the previous work, a tracking algorithm has been added into the loop: the classifier is used to detect the target only at the first step or in case of partial occlusions. Apart from such scenarios, a~\ac{CAMShift} tracking algorithm~\cite{Bradski1998} is employed to follow the car along the path, thus reducing the computational burden and the possibility to lose the target during the tracking. Moreover, in this paper it is proposed a novel procedure based on \textit{ad hoc} Matlab scripts that automatically select the bounding box area of the target (see, 
Fig.~\ref{fig:camshiftTrackingAlgorithm}) avoiding to use specific Matlab tools, such as \textit{Training Image Labeler}. These scripts also allow comparing various classifier configurations to help select the most suitable for the case study among different features types (e.g., Haar, HOG, LBP)~\cite{Rosten2006} and number of training stages. Finally, the software platform is published as open-source\footnote{\url{https://github.com/gsilano/MAT-Fly}} with the aim to share results with other researchers, students, and developers that might use the platform for testing their algorithms and understanding how different approaches can improve the performance and affect the system stability. 

The paper is organized as follows. Section~\ref{sec:systemDescription} explains the simulation scenario and its functionalities. The classifier training phase and the vision-based target detection and tracking algorithms are presented in Sec.~\ref{sec:classifierLearningPhase} and~\ref{sec:visionBasedTargetDetection}, respectively. Section~\ref{sec:modelOfAQuadrotorDrone} briefly describes the quad-rotor model while numerical results and the control algorithm are reported in Sec.~\ref{sec:flightControlSystem}. Finally, Section~\ref{sec:conclusion} concludes the 
paper.



\section{System Description}
\label{sec:systemDescription}

This section aims to describe MAT-Fly and how it works together with the Matlab~\ac{VR} and~\ac{CVS} toolboxes. An illustrative application, i.e., the \textit{object tracking example}, where a drone tracks a car moving along a nontrivial path is considered. An overview of the main elements that make up the system is depicted in Fig.~\ref{fig:systemSchematicDescription}. 

The software platform is mainly divided into four parts: the classifier training phase (see, Sec.~\ref{sec:classifierLearningPhase}), the vision-based target detection and tracking (see, Sec.~\ref{sec:visionBasedTargetDetection}), the flight control system (see, Sec.~\ref{sec:flightControlSystem}), and the Matlab~\ac{VR} toolbox. To facilitate the development of various control and computer vision strategies and the reuse of existing software components, the system was set up using a modular approach splitting each functionality into interchangeable modules. In other words, each part of the system (e.g., the vision-based target detection and tracking, the flight control system) was developed by isolating every feature (e.g., the detection algorithm, the reference generator) in such a way they can be easily replaced with others by facilitating the test and evaluation process.
\Figure[!t]()[width=0.48\textwidth]{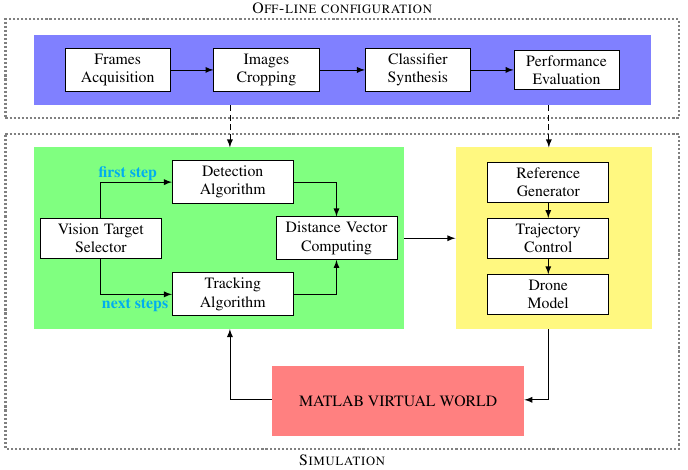}{The proposed software platform architecture. Arrows represent the data exchanged among blocks and how they interact with each other. Colors point out the four parts making up the system: the classifier training phase (in blue), the vision-based target detection and tracking (in green), the flight control system (in yellow), and the Matlab~\ac{VR} toolbox (in red). \label{fig:systemSchematicDescription}}

The Matlab~\ac{VR} toolbox allows to simulate a scenario as much similar as to the real world accounting for the interaction between complex dynamic systems with the surrounding scenario. Moreover, thanks to animation recording functionalities, frames and videos from the scene can be acquired and used to implement an~\ac{IBVS} problem. Also, the tool makes it easy to add external viewpoints to monitor any moving object in the 3D environment from different positions and orientations.
\Figure[!t]()[width=0.4\textwidth]{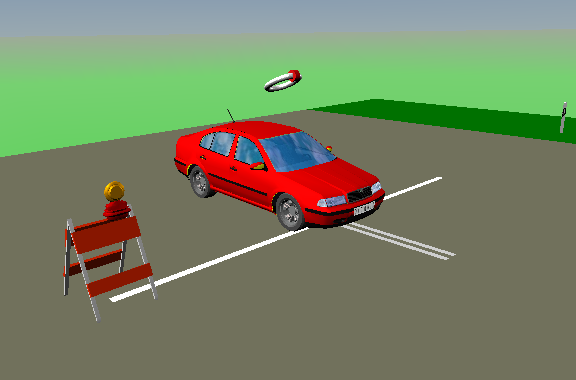} {Initial frame extracted from the \textit{object tracking example}. The steering angle visualizer allows monitoring the car movements along the path. \label{fig:initialFrameOfTheScenario}}

One of the available examples\footnote{The list of the ready-to-use scenarios is accessible at~\url{https://goo.gl/rtEx3S}.} (specifically the \textit{vr\_octavia\_2cars} example) that describes a quite detailed dynamical model of a car moving along a nontrivial path was used as a starting point (see, Fig.~\ref{fig:initialFrameOfTheScenario}). The example represents a standard double-lane-change maneuver~\cite{Arefnezhad2018IJAT} conducted in two-vehicles configuration, where one engages the~\ac{ESP} control while the other switches off such control unit when changing the lane. From this perspective, a simpler scenario was considered by removing one of the two vehicle configurations, i.e., the car without the~\ac{ESP} controller. 

Then, an external viewpoint was added to the scheme for simulating the behavior of a quad-rotor that flies by observing the car moving along the path. In Matlab~\ac{VR} a viewpoint has six~\acp{DOF}: the spatial coordinates $x$, $y$, and $z$, and the angles \textit{yaw} ($\psi$), \textit{pitch} ($\vartheta$), and \textit{roll} ($\varphi$). The whole process is the following: images are updated according to the position and the orientation of the quad-rotor w.r.t. the car; such images are acquired and elaborated for getting the necessary information to detect and track the target, and to run the control strategy designed for the tracking problem. The outputs of the control algorithm consists of the commands $u_\varphi$, $u_\vartheta$, $u_\psi$, and $u_T$ that should be given to the drone in order to update its position ($x_d$, $y_d$, and $z_d$) and orientation ($\varphi_d$, $\vartheta_d$, and $\psi_d$ ), see Fig.~\ref{fig:schemaDiControllo}. 

It is worth noticing that ground truth data are used by the tracking controller (see Sec.~\ref{sec:flightControlSystem}). Therefore, any analysis can be conducted on the correctness of the data and how this affects the navigation. However, this does not constitute a limitation for the proposed framework thanks to the modular interface exhibited by the 
platform~\cite{IMUMathWorks}.

In Figure~\ref{fig:simulinkScheme} the Simulink scheme employed for simulating the drone and the car dynamics is reported. The \textit{esp\_on} and the \textit{coordinates\_transformation} blocks compute the steering angle, the linear velocity and the position of the car, and all forces needed to follow a given path. Instead, the \textit{observer\_position} and \textit{rotation\_matrix} blocks represent the aircraft position and orientation (it is expressed by using the direction cosine matrix~\cite{Bouabdallah2005AR} and the Rodrigues's formula~\cite{Corke2011}), respectively. The processed data are sent to the \textit{VR Visualization} block that takes care of the drone and car movements in the simulated scenario. 
\Figure[!t]()[width=0.48\textwidth, height=1\textheight,keepaspectratio]{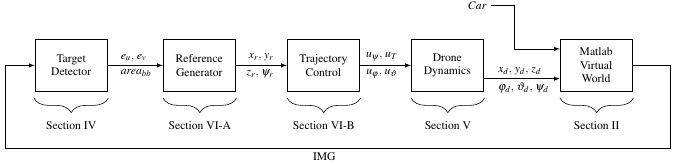}{The control scheme. Subscript \textit{d} indicates the drone variables, while \textit{r} represents the references to the controller. Each block is mapped with the section that describes its content to help matching the blocks with the corresponding description within the paper. \label{fig:schemaDiControllo}}

Note that Matlab VR adopts a reference system ($O_\mathrm{FVR}$)~\cite{MatlabVRUserGuide} slightly different from the classic fixed reference frame $O_\mathrm{FI}$ (see Sec.~\ref{sec:modelOfAQuadrotorDrone}), thus simple references transformations have been taken into account in all elaborations.

\Figure[!t]()[width=0.8\textwidth]{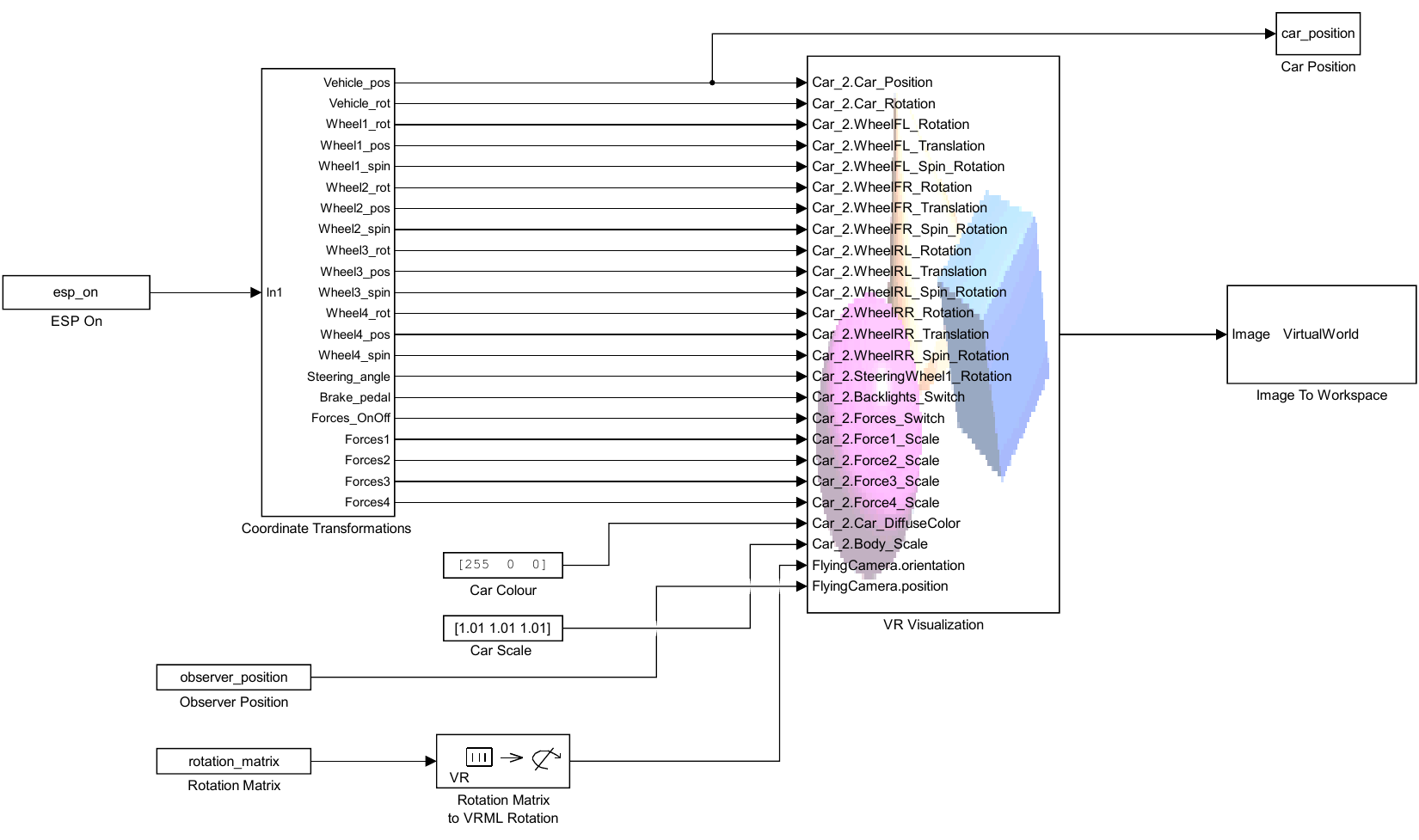}{Simulink scheme employed for simulating the drone and car dynamics in the 3D simulation environment.\label{fig:simulinkScheme}}

Finally, the Simulink scheme saves the current car position ($x_\mathrm{car}$, $y_\mathrm{car}$, and $z_\mathrm{car}$), used for comparing the drone and the car trajectories (see, Sec.~\ref{subsec:referenceGenerator}), and frames of the virtual scenario observed from the drone point of view. Those frames are used, as described in next sections, for pattern recognition.



\section{Classifier Training Phase}
\label{sec:classifierLearningPhase}

The classifier training phase is the most important part of the system: the object detection and tracking depend on it. Matlab scripts have been developed to automate the entire procedure, from the \textit{frames acquisition} to the \textit{classifier synthesis} and \textit{performance evaluation}. To this aim, the training process has been divided into four parts, as depicted in Fig.~\ref{fig:systemSchematicDescription}: the frames acquisition, the bounding box selection, the classifier synthesis and the performance evaluation.



\subsection{Frames acquisition}
\label{subsec:framesAcquisition}

When going to train a classifier, a high number of images is needed. The images are divided into two groups: \textit{positive} (that contain the target) and \textit{negative images}. Following what described in~\cite{Corke2011}, $2626$ positive and $10504$ negative images were used achieving a 1~:~4 ratio in accordance to the Pareto's principle (aka the $80/20$ rule).

For the frames acquisition, a simulation was performed with the quad-rotor moving along a spiral trajectory around the car parked in its initial state (see, Fig.~\ref{fig:droneTrajectoryAroundCar}). The aircraft attitude and position have been computed for each frame so as described by the sphere surface equations,
\Figure[!t]()[width=0.48\textwidth]{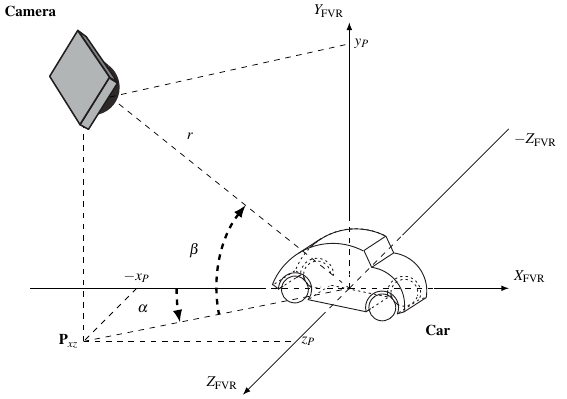}{Drone trajectory around the car parked in its initial state during the frames acquisition phase.\label{fig:droneTrajectoryAroundCar}}
\begin{equation}\label{eq:sfereSurface}
\left\{
\begin{array}{ll}
y=r\cos{\beta}\\
x=r\sin{\beta}\sin{\alpha}\\
z=r\sin{\beta}\cos{\alpha}
\end{array}
\right . ,
\end{equation}

where $r$, the sphere radius, is the distance between the car and the drone (assumed to be fixed and equal to $15$ meters), and together with $\alpha \in [0, 2\pi]$ and $\beta \in [0, \nicefrac{\pi}{2}]$ angles, identifies the drone position in the 3D space, as depicted in Fig.~\ref{fig:droneTrajectoryAroundCar}. A video showing the quad-rotor camera point of view while observing the car parked in its initial state while following the spiral trajectory is available in~\cite{Silano_FramesAcquire2017}.



\subsection{Bounding box selection}
\label{subsec:imageCropping}

To train the classifier, the~\ac{ROI} of the target needs to be computed. Due to the high number of images, manual labeling tools, such as the MathWorks \textit{Training Image Labeler}, cannot be used. Thus a Matlab script was developed to automatically select the bounding box area surrounding the target. The image segmentation process was used to simplify and to change the image representation: from RGB to grayscale (Figs.~7(a) and~7(b), respectively). The result is a set of contours that make the image meaningful and easier to analyze: each group of pixels in a region is similar w.r.t. some characteristics or computed properties, intensity, or texture, while adjacent regions are significantly different w.r.t. the same properties.
\Figure[!t]()[width=0.48\textwidth]{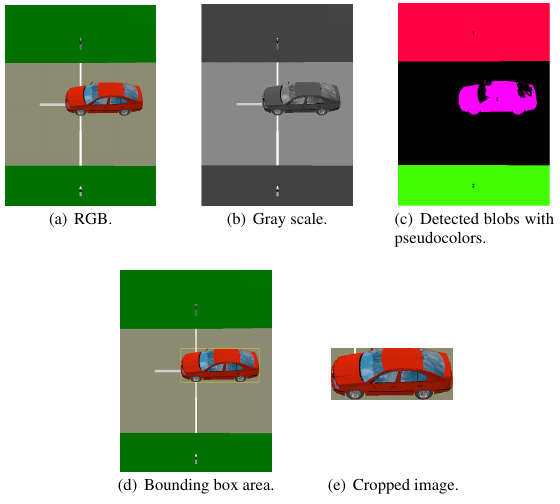}{Frames obtained by the images segmentation process. From the RGB file format (a) to the cropped image (e) the steps are shown, sequentially.\label{fig:imageSegmentationSteps}}

To automatically select each group of pixels, the~\ac{BTH} method~\cite{DosAnjos2008} was used. Such a method allows to separate the background from the foreground image by dividing the data into two main classes (see, Fig.~\ref{fig:histogram}) and by searching for the optimum threshold level. 
\Figure[!t]()[width=0.35\textwidth]{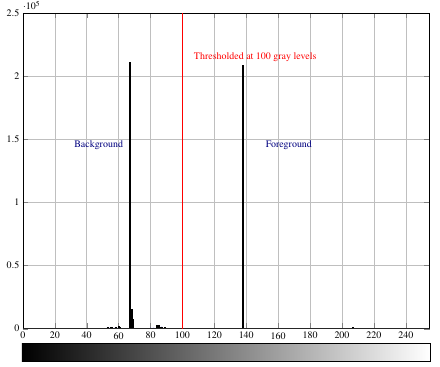}{Histogram of the image data. The red line, i.e., the grayscale threshold, divides the graph into two parts: the \textit{background} and the \textit{foreground}. The gray gradient bars indicate the associated color to each $x$-value, from $0$ to $255$.\label{fig:histogram}}

Starting from the foreground grayscale image (see, Fig.~7(b)), the script deals with labeling the individual blobs by using the connected-component labeling algorithm~\cite{Corke2011} with a fixed heuristic (8-connected, in the considered case). In Figure~7(c) the obtained blobs are depicted with different colors and numbers for visual convenience. Then, the script selects the blob that meets the criteria in terms of size and intensity (chosen to match the target properties) in order to obtain a unique bounding box surrounding the target (see, Figs.~7(d) and~7(e)). Finally, the script provides as output a MAT-file containing, for each positive image, the suitable~\ac{ROI} components, i.e., the bounding box centroid, its width and height. This file is used for the classifier synthesis in the target detection design process (see, Fig.~\ref{fig:systemSchematicDescription}).

The proposed approach allows to automatically label the target (the car) from the positive images, thus decreasing the time spent for the training phase. In Figure~\ref{fig:imageSegmentationSteps}, for a single sample frame, all elaboration steps are reported.

On the considered data set, the script was able to automatically detect the~\acp{ROI} with an error of $\SI{8.18}{\percent}$: the target was not recognized only in $215$ frames out of $2626$ positive images, and the first~\ac{ROI} loss appeared at the $1791$\textsuperscript{th} frame.



\subsection{Classifier synthesis}
\label{subsec:classifierSynthesis}

The Viola \& Jones algorithm~\cite{Viola2001} was chosen as object detection framework to recognize the car along the path. The algorithm was originally designed and developed for face detection problem, but it can be easily trained to detect any object~\cite{Xu2017} by using different features types (e.g., Haar, HOG, LBP)~\cite{Rosten2006} and training stages\footnote{This is common in cascade classifiers where each stage is an ensemble of weak learners, i.e., simple classifiers called decision stumps.}. Although even more complicated and performing object detection frameworks (e.g., YOLO~\cite{Redmon2016CVPR, Jiao2019Access}, Fast R-CNN~\cite{Girshick2015ICCV, Jiao2019Access}) are available in the literature, historical reasons motivated this choice: the Viola \& Jones classifier was the first object detection framework in real-time. Thus, it is of interest, at least for educational purposes, to have a simulation platform that performs object detection with such a solution.

For the considered testbed, the Haar features were used to design the classifier. These features along with LBP are often used to detect faces due to their fine-scale textures while HOG features are often employed to detect objects. However, the obtained results suggested to choose Haar features which appeared more useful for capturing the overall shape of the target (see, Fig.~\ref{fig:detectorComparison}) even if longer time was needed during the training phase.
\Figure[!t]()[width=0.48\textwidth]{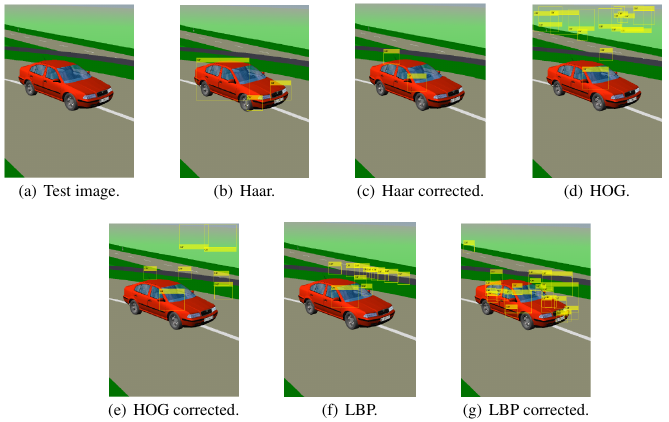}{Detection results obtained by using Haar, HOG and LBP feature types. The false alarm and true positive rates are $0.001$ and $0.995$, respectively, while the number of training stages is $4$. Two different target models are considered: the ``corrected'' and ``uncorrected'' version. \label{fig:detectorComparison}}



\subsection{Performance comparison}
\label{subsec:performanceComparison}

When designing a cascade object detector, the number of training stages, the false alarm and true positive rates, need to be tuned in accordance to the required performance (e.g., accuracy) and constraints (e.g., time response). To facilitate the analysis as well as to find the most suitable set of parameters that fit the problem, a Matlab script was developed to evaluate the performance of the classifier. This script is part of the proposed software platform (see, 
Fig.~\ref{fig:systemSchematicDescription}) and allows to compare in a few steps various configurations and models getting a general overview of how the object detector behaves. 

In Fig.~\ref{fig:detectorComparison} the results obtained for a single sample frame are reported. Two different models were considered to prove the validity of the proposed approach: the \textit{uncorrected} and \textit{corrected} models. The first one uses the~\acp{ROI} automatically extracted from the algorithm presented in Sec.~\ref{subsec:imageCropping}, while the second one employs those obtained using the Matlab tool \textit{Training Image Labeler}. In all revelations, the car is only partially detected despite the large number of images employed to train the classifier. Except for some cases, there are no revelation errors: different bounding boxes are detected in the image. This is probably due to the absence of photorealism in the collected frames. As described in~\cite{Gonzalez2019}, the reality gap (i.e., realistic geometry, textures, lighting conditions, camera noise, and distortion) affects the performance of computer vision algorithms. On the other hand, they introduce enough ``useful noise'' to help the detection (the presence of several image view points). 

Many tests have been conducted in order to assess the true performance of the classifier. As shown in Figs.~9(b) and~9(c), the detection results are very similar for both models. Thus, it is a good approximation to consider the ``uncorrected''~\acp{ROI} instead of the ``corrected'' version in the classifier design process. Such approximation allows to save time during the training phase thus avoiding to use specific tools, such as the Matlab tool \textit{Training Image Labeler}, when the~\ac{ROI} detection fails. Moreover, it proves the validity and the effectiveness of the automatic tool 
procedure for bounding box selection. Of course, further tests may be carried out considering more valuable evaluation criteria, such as confusion matrix, accuracy, precision, recall, specificity~\cite{Jiao2019Access}. 



\section{Vision-Based Target Detection}
\label{sec:visionBasedTargetDetection}

The vision-based target detection phase sets up the~\ac{IBVS} problem using the classifier and tracking algorithms as feedback from the environment (see, Fig.~\ref{fig:systemSchematicDescription}). There are four components that constitute this part: the \textit{vision target selector}, the \textit{detection} and \textit{tracking algorithm}, and the \textit{distance vector computing}. 
\Figure[!t]()[width=0.48\textwidth]{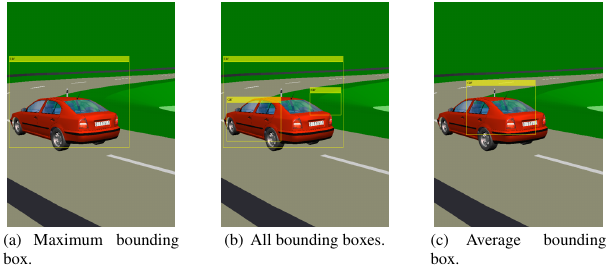}{Bounding box selection algorithm. The detection results are obtained by using the Haar cascade features type. The maximum (left) and average (right) bounding boxes are computed by using the result obtained from detection (center). \label{fig:maximumAverageBoundingBox}}

The vision target selector takes care of switching between the detection and the tracking algorithms based on the recognition results: the detector is used only at the first step or in case of partial occlusion, otherwise a~\ac{CAMShift} tracking algorithm~\cite{Corke2011} is employed to follow the car along the path. This algorithm performs target tracking by searching for its probability distribution pattern in a local adaptive size window. Although it does not guarantee 
the best performances, the algorithm supplies reliable and robust results~\cite{Artner2008}. 

Due to multiple target revelations (see, Sec.~\ref{subsec:performanceComparison}), a Matlab script was used to obtain a unique bounding box surrounding the target (the car). The script computes the maximum (Fig.~10(a)) and the average (Fig.~10(c)) bounding boxes, as shown in Fig.~\ref{fig:maximumAverageBoundingBox}. The maximum approach puts more trust in the detection results, while the average approach tries to filter out the revelation errors. The ``good'' choice depends on the particular employed classifier and on the amount of frames used during the training phase. For the considered testbed, the maximum bounding box was chosen to figure out the~\ac{IBVS} problem. 

Once the target has been recognized, the distance vector computing block generates the references for the drone trajectory control (see, Sec.~\ref{sec:flightControlSystem} and Fig.~\ref{fig:schemaDiControllo}) measuring the distance between the image ($u_\mathrm{img}$, $v_\mathrm{img}$) and bounding box ($u_{bb}$, $v_{bb}$) centroids, as depicted in 
Fig.~\ref{fig:boundingBoxOttenutaDetection}. The vector aims to provide the reference signals ($e_u$ and $e_v$) to move the drone so that the center of the bounding box surrounding the car overlaps the centroid of the image.
\Figure[!t]()[width=0.48\textwidth]{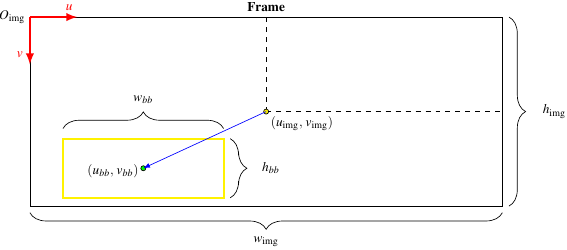}{The diagram shows how frames are processed once the target is detected. The distance vector and the bounding box are represented in blue and yellow, respectively. The image ($u_\mathrm{img}$, $v_\mathrm{img}$) and bounding box ($u_{bb}$, $v_{bb}$) centroids are also reported. \label{fig:boundingBoxOttenutaDetection}}




\section{Drone Dynamical Model}
\label{sec:modelOfAQuadrotorDrone}

For the specific case study, a quad-rotor in a plus configuration has been considered. The design of a high performance attitude and position controller requires often an accurate model of the system. It is here recalled the commonly used dynamical model of a quad-rotor~\cite{Bouabdallah2005ICRA} and, by following usual approaches, two orthonormal frames are introduced: the fixed-frame $O_\mathrm{FI}$ (where FI stands for Fixed Inertial), also called inertial (or reference) frame, and the body-frame $O_\mathrm{ABC}$ (where ABC stands for Aircraft Body Center) that is fixed in the aircraft center of mass and is oriented according to the aircraft orientation, see Fig.~\ref{fig:droneModel}.
\Figure[!t]()[width=0.4\textwidth]{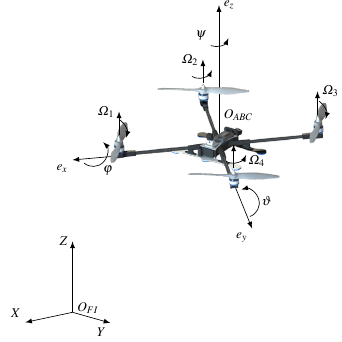}{Drone in the body-frame ($O_\mathrm{ABC}$) and the fixed-frame ($O_\mathrm{FI}$) reference systems. Forces produced by each rotor, spin directions and propeller velocities, $\Omega_i$, are also reported.\label{fig:droneModel}}

The translational dynamic equations of the aircraft can be expressed in the inertial frame as follows:
\begin{align}\label{eq:translationalDynamics}
m \pmb{\ddot{\xi}} = - m g \mathbf{E}_z + u_T \mathbf{R}(\varphi,\vartheta,\psi) \mathbf{E}_z,
\end{align}
where $g$ denotes the gravity acceleration, $m$ the mass, $u_T$ the total thrust produced by the rotors, $\pmb{\xi}=(x, y, z)^\top \in \mathbb{R}^3$ the drone position expressed in the inertial frame, $\mathbf{E}_z=(0, 0, 1)^\top$ is the unit vector along the $Z$-axis, while $\mathbf{R}(\varphi,\vartheta,\psi) \in \mathbb{R}^{3 \times 3}$ is the rotation matrix from the body to the inertial frame and it depends on the attitude $\pmb{\eta}=(\varphi, \vartheta, \psi)^\top \in \mathbb{R}^3$ (i.e., Euler angles roll, pitch, and yaw, respectively) that describes the body-frame orientation according to the ZYX convention~\cite{Stevens}. Furthermore, the rotational dynamics can be expressed as
\begin{equation}\label{eq:rotationalDynamics}
\mathbf{I} \pmb{\dot{\mathbf{\omega}}}_B = - \pmb{\omega}_B \times \mathbf{I} 
\pmb{\mathbf{\omega}}_B + \pmb{\tau},
\end{equation}
where `$\times$' denotes the vector product, $\pmb{\omega}_B=(\omega_x, \omega_y, \omega_z)^\top \in \mathbb{R}^3$ is the angular velocity vector expressed in the body-frame, $\mathbf{I} = \text{diag}(I_x,\,I_y,\,I_z) \in \mathbb{R}^{3 \times 3}$ is the inertia matrix of the vehicle w.r.t. its principal axes, and $\pmb{\tau}=(u_{\varphi}, u_{\vartheta}, u_{\psi})^\top \in \mathbb{R}^3$ is the control torque vector obtained by actuating the rotors speeds according to the 
rotors configuration and the vehicle shape.

At low speeds and around the hovering state, the simplified dynamic model consists of six second order differential equations obtained from balancing forces  and momenta acting on the drone, where $c_\bullet$ and $s_\bullet$ denote the $\cos(\bullet)$ and $\sin(\bullet)$ functions, respectively:  
\begin{subequations}\label{eq:rotationalSubsystem} 
	\begin{align}
	I_x \ddot \varphi = & \dot \vartheta \dot \psi  \left(I_y - I_z\right) + u_{\varphi}, \\
	I_y \ddot \vartheta  = &\dot \varphi \dot \psi  \left(I_z - I_x\right) + u_{\vartheta},\\
	I_z \ddot \psi  = & \dot \vartheta \dot \varphi  \left(I_x - I_y\right) + u_{\psi} ,
	\end{align}
\end{subequations}
\vspace{-0.7cm}
\begin{subequations}\label{eq:translationalSubystem}
	\begin{align}
	m\ddot{x} = & u_T \left(c_{\varphi} s_{\vartheta} c_{\psi} + s_{\varphi} 
	s_{\psi}\right), \\
	m\ddot{y} = & u_T \left(c_{\varphi} s_{\vartheta} s_{\psi} - s_{\varphi} 
	c_{\psi}\right), \\
	m\ddot{z} = & u_T c_{\vartheta} c_{\varphi} - mg ,
	\end{align}	
\end{subequations}
%
with
\begin{align}\label{eq:totalThrust}
u_T = b_f\left(\Omega_1^2 + \Omega_2^2 + \Omega_3^2 + \Omega_4^2\right),
\end{align}
and
\begin{align}\label{eq:ctrlMixer}
\begin{pmatrix}
u_\varphi   \\
u_\vartheta \\
u_\psi  
\end{pmatrix}
= 
\dfrac{b_f}{b_m}
\begin{pmatrix}
b_m l \left( \Omega_4^2 - \Omega_2^2\right)\\
b_m l \left( \Omega_3^2 - \Omega_1^2\right) \\
- \Omega_1^2 + \Omega_2^2 - \Omega_3^2 + \Omega_4^2 	
\end{pmatrix},
\end{align}
where $\Omega_i$, $i~\in~\{1,2,3,4\}$, are the actual rotors angular velocities expressed in \si{\radian\per\second}, $l$ is the	distance from the propellers to the center of mass, while $b_f$ and $b_m$ are the thrust and drag factors, respectively. Further details can be found in~\cite{Bouabdallah2005ICRA, Bouabdallah2005AR, Stevens}. Table~\ref{tab:systemParameters} reports the parameters values of the drone for the considered case study (see Sec.~\ref{subsec:numericalResults}).
\begin{table}
	\centering
	\begin{tabular}{|l|c|c|c|}
		\hline
		& \textbf{Sym.} & \textbf{Value} & \textbf{Unit}\\
		\hline
		\hline
		Mass & $m$ & $0.65$ & $\si{\kilogram}$\\
		Distance to center of gravity & $l$ & $0.23$ & $\si{\meter}$\\ 
		Thrust factor & $b_f$ & $7.5 \cdot 10^{-7}$ & $\si{\kilogram}$\\
		Drag factor & $b_m$ & $3.13 \cdot 10^{-5}$ & $\si{\kilogram\meter}$\\ 
		Inertia component along $e_x$-axis & $I_x$ & $7.5 \cdot 10^{-3}$ & 
		$\si{\kilogram\meter\squared}$\\
		Inertia component along $e_y$-axis & $I_y$ & $7.5 \cdot 10^{-3}$ & 
		$\si{\kilogram\meter\squared}$\\
		Inertia component along $e_z$-axis & $I_z$ & $1.3 \cdot 10^{-3}$ & 
		$\si{\kilogram\meter\squared}$\\
		\hline
	\end{tabular}
	\caption{Drone parameter values for the considered case study.}
	\label{tab:systemParameters}
\end{table}



\section{Flight Control System}
\label{sec:flightControlSystem}

Various state-of-the-art solutions investigate the trajectory tracking problem with quad-rotors. However, not all of them are suitable for the specific case of application~\cite{Nascimento2019}. Therefore, with the aim of illustrating a control design methodology exploiting the~\ac{IBVS} approach, it has been considered the flight control system described in~\cite{Bouabdallah2005AR} and~\cite{Pestana2014} that uses a \textit{reference generator} and an \textit{integral 
backstepping} \textit{controller} to figure out the drone trajectory tracking problem. The reference generator extracts the information from the images to generate the path to follow, while the~\ac{IB} controller uses those references to compute the needed drone command signals. Figures~\ref{fig:referenceGeneratorScheme} and~\ref{fig:schemaControlloAssettoPID} describe the overall control scheme. 
\Figure[!t]()[width=0.475\textwidth]{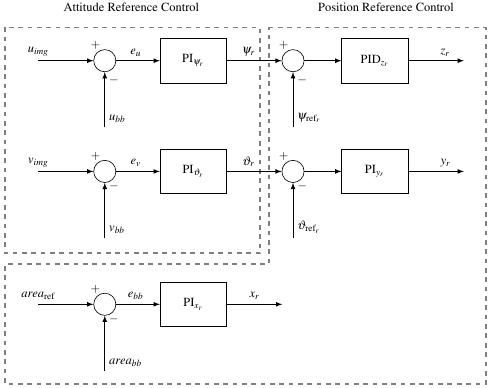}{The reference generator scheme referred to virtual reference system ($O_\mathrm{FVR}$). The obtained heuristic PID gains are: $K_{P_{\psi_r}}=1\cdot 10^{-5}$, $K_{I_{\psi_r}}=1\cdot 10^{-3}$, $K_{P_{\vartheta_r}}=1\cdot 10^{-5}$, $K_{I_{\vartheta_r}}=1\cdot 10^{-3}$, 
$K_{P_{x_r}}=1\cdot10^{-6}$, $K_{I_{x_r}}=6\cdot10^{-6}$, $K_{P_{y_r}}=1\cdot10^{-2}$, 
$K_{I_{y_r}}=1\cdot10^{-2}$, $K_{P_{z_r}}=15$, $K_{I_{z_r}}=57.5$ and $K_{D_{z_r}}=3.75$. \label{fig:referenceGeneratorScheme}}
\Figure[!t]()[width=0.47\textwidth]{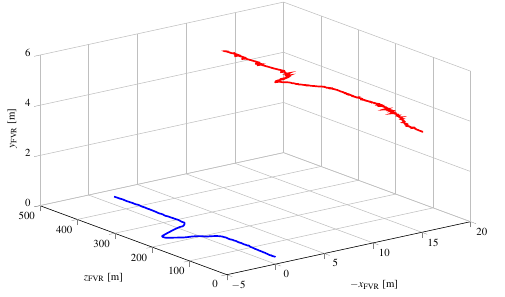}{The car (in blue) and the output of the reference generator (in red) while tracking the target.\label{fig:pathTargetFollowing}}



\subsection{Reference Generator}
\label{subsec:referenceGenerator}

The reference generator is decomposed into two parts: the attitude and the position controller, both illustrated in Fig.~\ref{fig:referenceGeneratorScheme}. The attitude controller tunes the yaw ($\psi_r$) and the pitch ($\vartheta_r$) angles trying to overlap the image ($u_\mathrm{img}$, $v_\mathrm{img}$) and the bounding box ($u_{bb}$, $v_{bb}$) centroids (see, Fig.~\ref{fig:boundingBoxOttenutaDetection}), while the roll ($\varphi_r$) angle is computed by the IB controller\footnote{All elaborations are expressed in the $O_\mathrm{FVR}$ reference system.}. These values are later used by the position controller to vary the drone reference position $z_r$ and $y_r$, while $x_r$ is computed comparing the detected area $area_\mathrm{bb}$ with those obtained when training the classifier $area_\mathrm{ref}$\footnote{This values is obtained as sample of mean of the collected~\acp{ROI} while training the classifier (see Sec.~\ref{subsec:framesAcquisition}).}

The proposed control architecture is based on control loops that are nothing but~\ac{PID} controllers. These are a standard solution in the literature for quad-rotor control design~\cite{Dief2015}. For the considered case study, the vehicle starts flying $4$ meters over the ground ($Z$-axis) with a distance of $15$ meters from the car along the $X$-axis in the $O_\mathrm{FVR}$ reference system.

Figure~\ref{fig:pathTargetFollowing} reports the trajectories followed by the car and the drone when running the simulation, while a further video has been made available at~\cite{Silano2017_referenceGenerator}. 



\subsection{Integral Backstepping controller}
\label{subsec:integralBacksteppeginController}

The integral backstepping of~\cite{Bouabdallah2005AR, Bouabdallah2005ICRA} has been used as trajectory controller for the path tracking. It performs robustness against external disturbances (offered by backstepping) and sturdiness w.r.t. model uncertainties (given by the integral action). Starting from the outputs of the reference generator, the~\ac{IB} controller computes the orientation  ($\varphi_{\mathrm{ref}_\mathrm{IB}}$ and $\vartheta_{\mathrm{ref}_\mathrm{IB}}$) that the drone 
should assume to follow the reference path ($x_r$ and $z_r$). The $\varphi_{\mathrm{ref}_\mathrm{IB}}$ and $\vartheta_{\mathrm{ref}_\mathrm{IB}}$ reference angles are computed as:
\begin{subequations}
	\begin{align}
	\vartheta_{\mathrm{ref}_\mathrm{IB}}=&\dfrac{m}{u_T} \Biggl[ \left ( 1-c_1^2+\lambda_1 \right 
	)e_x+\left(c_1+c_{2}\right)e_{x_\mathrm{IB}}+\\&-c_1 \lambda_1 \int_0^t{e_x 
	(\tau)\mathrm{d}\tau} \Biggr] \nonumber,
	\end{align}
\end{subequations}

\begin{subequations}
\begin{align}
	\varphi_{\mathrm{ref}_\mathrm{IB}}=&-\cfrac{m}{u_T} \Biggl[\left (1-c_{3}^2+\lambda_2 \right 
	)e_z+\left(c_{3}+c_{4}\right)e_{z_\mathrm{IB}}+\\&-c_{3}\lambda_2 \int_0^t{e_z 
	(\tau)\mathrm{d}\tau} \Biggr] \nonumber,
	\end{align}
\end{subequations}
with
\begin{subequations}
	\begin{align}
	e_{x_\mathrm{IB}}(t)&=\lambda_1 \int_0^t{e_x (\tau)\mathrm{d}\tau}+c_1 e_x(t)+\dot{e}_x(t), \\
	e_{z_\mathrm{IB}}(t)&=\lambda_2 \int_0^t{e_z (\tau)\mathrm{d}\tau}+c_{3} e_z(t)+\dot{e}_z(t), 
	\end{align}
\end{subequations}
and
\begin{subequations}
	\begin{align}
	e_x&=x_r-x_d, \\
	e_z&=z_r-z_d .
	\end{align}
\end{subequations}
For the considered motivating examples, the following values have been chosen: $\lambda_1=0.025$, $\lambda_2=0.025$, $c_1=2$, $c_2=0.5$, $c_3=2$ and $c_4=0.5$. Figure~\ref{fig:schemaControlloAssettoPID} shows the overall control system architecture.
\Figure[!t]()[width=0.42\textwidth]{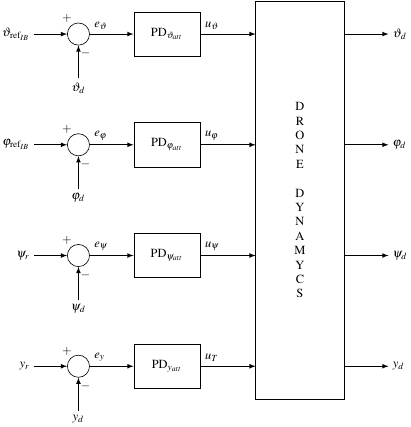}{The drone trajectory controller. All variables are expressed in the virtual reference system ($O_\mathrm{FVR}$). Here we recall the heuristic control gains employed into the simulation scenario: $K_{P_{y_{att}}}=1000$, $K_{D_{y_{att}}}=200$, $K_{P_{\varphi_{att}}}=8$, $K_{D_{\varphi_{att}}}=4$, $K_{P_{\vartheta_{att}}}=12$, $K_{D_{\vartheta_{att}}}=4$, $K_{P_{\psi_{att}}}=10$, and $K_{D_{\psi_{att}}}=4$. \label{fig:schemaControlloAssettoPID}}



\subsection{Numerical results}
\label{subsec:numericalResults}

To prove the validity and effectiveness of the proposed framework, numerical simulations have been carried out by using the 2015b release of Matlab equipped with~\ac{CVS} and~\ac{VR} toolboxes\footnote{The simulator is fully compatible with each further release of Matlab.}. The video available at~\cite{Silano2017_droneFollowsACar} illustrates in a direct way how the system works, i.e., the ability of the quad-rotor to follow the car that moves along the nontrivial path. In addition, the video shows the behavior of the detection and tracking algorithms that never lose the target while tracking the target. Moreover, the video shows the capabilities of the control system in reacting to changes in the car's dynamics: during the double lane change maneuver the vehicle suddenly increases its speed and the aircraft tilts around the $X$-axis (the car seems to climb a hill\footnote{The drone flies in an eye-in-hand configuration, i.e., tilts around the axis direct affects the camera orientation.}) to capture the shift in the dynamics.

A further scenario (the video is available at~\cite{Silano2017_carPartiallyCover}) was considered to show how the simulation can be easily customized without the need to redesign the entire system. The numerical example aims to show how the performance of the detection and tracking algorithms can be easily evaluated while running alongside the drone tracking controller. Two cars are considered for the case of interest. One (red car) engages the~\ac{ESP} control while the other (yellow) switches off such control unit when changing the lane. As it can be seen from the video, the search window
adapts its sizes in response to partial occlusions of the target.

Finally, the video at~\cite{worldChanged} shows the advantages of using a modular architecture for the platform. The video shows how the whole system architecture can be tested under various conditions simply changing the scenario\footnote{The \textit{vr\_octavia} scenario was considered.}. As shown in the video, halfway through the simulation (\SI{26}{\second}) increasing the speed of the car causes the drone to tilt excessively moving to instability.

The proposed scenarios demonstrate as the software platform allows to test the complex system while interacting with the surrounding environment and computer vision and control algorithms are in the loop.



\section{Conclusions}
\label{sec:conclusion}

In this paper, a numerical simulation platform for multi-rotor aircraft based on Matlab and the MathWorks Virtual Reality and Computer Vision System toolboxes has been described. The platform makes easy to implement and to simulate complex scenarios where computer vision algorithms can be run and tested together with drone tracking controllers. The simulator provides a ready-to-use environment allowing students, researchers, and developers to easily test and evaluate their own algorithms. The platform also constitutes the first step towards the development of a more structured software tool where exploiting the advantages of software-in-the-loop simulations. The software has been released as open-source\textsuperscript{3} making it possible to go through any part of the system. Future work includes the integration of the platform with more advanced robotics middleware and the creation of the interface with the hardware moving toward hardware-in-the-loop tests.



\bibliographystyle{IEEEtran}
\bibliography{bib}

\begin{thebibliography}{10}
\providecommand{\url}[1]{#1}
\csname url@rmstyle\endcsname
\providecommand{\newblock}{\relax}
\providecommand{\bibinfo}[2]{#2}
\providecommand\BIBentrySTDinterwordspacing{\spaceskip=0pt\relax}
\providecommand\BIBentryALTinterwordstretchfactor{4}
\providecommand\BIBentryALTinterwordspacing{\spaceskip=\fontdimen2\font plus
\BIBentryALTinterwordstretchfactor\fontdimen3\font minus
  \fontdimen4\font\relax}
\providecommand\BIBforeignlanguage[2]{{%
\expandafter\ifx\csname l@#1\endcsname\relax
\typeout{** WARNING: IEEEtran.bst: No hyphenation pattern has been}%
\typeout{** loaded for the language `#1'. Using the pattern for}%
\typeout{** the default language instead.}%
\else
\language=\csname l@#1\endcsname
\fi
#2}}

\bibitem{Scaramuzza2014}
D.~{Scaramuzza}, M.~C. {Achtelik}, L.~{Doitsidis}, F.~{Friedrich},
  E.~{Kosmatopoulos}, A.~{Martinelli}, M.~W. {Achtelik}, M.~{Chli},
  S.~{Chatzichristofis}, L.~{Kneip}, D.~{Gurdan}, L.~{Heng}, G.~H. {Lee},
  S.~{Lynen}, M.~{Pollefeys}, A.~{Renzaglia}, R.~{Siegwart}, J.~C. {Stumpf},
  P.~{Tanskanen}, C.~{Troiani}, S.~{Weiss}, and L.~{Meier},
  ``{Vision-Controlled Micro Flying Robots: From System Design to Autonomous
  Navigation and Mapping in GPS-Denied Environments},'' \emph{IEEE Robotics
  Automation Magazine}, vol.~21, no.~3, pp. 26--40, 2014.

\bibitem{Choi2015}
S.~Choi and E.~Kim, ``{Image acquisition system for construction inspection
  based on small unmanned aerial vehicle},'' \emph{Lecture Notes in Electrical
  Engineering}, vol. 352, pp. 273--280, 2015.

\bibitem{Fraundorfer2012}
F.~{Fraundorfer}, L.~{Heng}, D.~{Honegger}, G.~H. {Lee}, L.~{Meier},
  P.~{Tanskanen}, and M.~{Pollefeys}, ``{Vision-based autonomous mapping and
  exploration using a quadrotor},'' in \emph{IEEE International Conference on
  Intelligent Robots and Systems}, 2012, pp. 4557--4564.

\bibitem{Anthony2014}
D.~{Anthony}, S.~{Elbaum}, A.~{Lorenz}, and C.~{Detweiler}, ``{On crop height
  estimation with UAVs},'' in \emph{IEEE International Conference on
  Intelligent Robots and Systems}, 2014, pp. 4805--4812.

\bibitem{Blosch2010}
M.~{Blosch}, S.~{Weiss}, D.~{Scaramuzza}, and R.~{Siegwart}, ``{Vision based
  MAV navigation in unknown and unstructured environments},'' in \emph{IEEE
  International Conference on Robotics and Automation}, 2010, pp. 21--28.

\bibitem{Castro2016}
D.~{Ferreira de Castro} and D.~A. {dos Santos}, ``{A Software-in-the-Loop
  Simulation Scheme for Position Formation Flight of Multicopters},''
  \emph{Journal of Aerospace Technology and Management}, vol.~8, no.~4, pp.
  431--440, 2016.

\bibitem{Mancini2017}
M.~{Mancini}, G.~{Costante}, P.~{Valigi}, T.~A. {Ciarfuglia}, J.~{Delmerico},
  and D.~{Scaramuzza}, ``{Toward Domain Independence for Learning-Based
  Monocular Depth Estimation},'' \emph{IEEE Robotics and Automation Letters},
  vol.~2, no.~3, pp. 1778--1785, 2017.

\bibitem{Abhijeet2015}
A.~{Tallavajhula} and A.~{Kelly}, ``{Construction and validation of a high
  delity simulator for a planar range sensor},'' in \emph{IEEE Conference on
  Robotics and Automation}, 2015, pp. 1050--4729.

\bibitem{Rosen2008}
R.~{Dianko} and J.~{Kuffner}, ``{Openrave: A planning architecture for
  autonomous robotics},'' Robotics Institute, Pittsburgh, PA, Tech. Rep.~79,
  2008.

\bibitem{Silano2020SprinerBook}
G.~{Silano} and L.~{Iannelli}, ``{CrazyS: A Software-in-the-Loop Simulation
  Platform for the Crazyflie 2.0 Nano-Quadcopter},'' in \emph{Robot Operating
  System (ROS): The Complete Reference (Volume 4)}, {Koubaa, Anis}, Ed.\hskip
  1em plus 0.5em minus 0.4em\relax Springer International Publishing, 2020, pp.
  81--115.

\bibitem{Sinha2021SMPT}
S.~{Sinha}, N.~K. {Goyal}, and R.~{Mall}, ``{Reliability and availability
  prediction of embedded systems based on environment modeling and
  simulation},'' \emph{{Simulation Modelling Practice and Theory}}, vol. 108,
  pp. 1--26, 2021.

\bibitem{Silano2019SMC}
G.~{Silano}, P.~{Oppido}, and L.~{Iannelli}, ``{Software-in-the-loop simulation
  for improving flight control system design: a quadrotor case study},'' in
  \emph{IEEE International Conference on Systems, Man and Cybernetics}, 2019,
  pp. 466--471.

\bibitem{Elkady2012}
A.~Elkady and T.~Sobh, ``{Robotics Middleware: A Comprehensive Literature
  Survey and Attribute-Based Bibliography},'' \emph{Journal of Robotics}, 2012.

\bibitem{Koenig2004}
N.~{Koenig} and A.~{Howard}, ``{Design and use paradigms for Gazebo, an
  open-source multi-robot simulator},'' in \emph{Proceedings of the IEEE
  International Conference on Intelligent Robots and Systems}, vol.~3, 2004,
  pp. 2149--2154.

\bibitem{VREP2013}
E.~{Rohmer}, S.~P.~N. {Singh}, and M.~{Freese}, ``{V-REP: a Versatile and
  Scalable Robot Simulation Framework},'' in \emph{Proceedings of The
  International Conference on Intelligent Robots and Systems}, 2013, pp.
  1321--1326.

\bibitem{airsim2017fsr}
S.~{Shah}, D.~{Dey}, C.~{Lovett}, and A.~{Kapoor}, ``{AirSim: High-Fidelity
  Visual and Physical Simulation for Autonomous Vehicles},'' in \emph{Field and
  Service Robotics}, M.~{Hutter} and R.~{Siegwart}, Eds.\hskip 1em plus 0.5em
  minus 0.4em\relax Springer International Publishing, 2018, pp. 621--635.

\bibitem{Echeverria2011}
G.~Echeverria, N.~Lassabe, A.~Degroote, and S.~Lemaignan, ``{Modular open
  robots simulation engine: {MORSE}},'' in \emph{IEEE International Conference
  on Robotics and Automation}, 2011, pp. 46--51.

\bibitem{Quigley2009}
M.~Quigley, K.~Conley, B.~Gerkey, J.~Faust, T.~Foote, J.~Leibs, R.~Wheeler, and
  A.~Y. Ng, ``{ROS: an open-source Robot Operating System},'' in
  \emph{Proceedings ICRA Workshop Open Source Software}, 2009, pp. 1--6.

\bibitem{Metta2006}
G.~Metta, P.~Fitzpatrick, and L.~Natale, ``{YARP: Yet another robot
  platform},'' \emph{International Journal on Advanced Robotics Systems},
  vol.~3, no.~1, pp. 43--48, 2006.

\bibitem{Mallet2002}
A.~Mallet, S.~Fleury, and H.~Bruyninckx, ``{A specification of generic robotics
  software components: future evolutions of GenoM in the Orocos context},'' in
  \emph{IEEE International Conference on Intelligent Robots and Systems}, 2002,
  pp. 2292--2297.

\bibitem{Khan2017}
S.~Khan, M.~H. Jaffery, A.~Hanif, and M.~R. Asif, ``{Teaching Tool for a
  Control Systems Laboratory Using a Quadrotor as a Plant in MATLAB},''
  \emph{IEEE Transactions on Education}, vol.~60, no.~99, pp. 1--8, 2017.

\bibitem{Day2015}
M.~A. {Day}, M.~R. {Clement}, J.~D. {Russo}, D.~{Davis}, and T.~H. {Chung},
  ``{Multi-uav software systems and simulation architecture},'' in
  \emph{International Conference on Unmanned Aircraft Systems}, 2015, pp.
  426--435.

\bibitem{Shokry2009}
H.~{Shokry} and M.~{Hinchey}, ``{Model-Based Verification of Embedded
  Software},'' \emph{Computer}, vol.~42, no.~4, pp. 53--59, 2009.

\bibitem{Unicomb2017IROS}
J.~{Unicomb}, L.~{Dantanarayana}, J.~{Arukgoda}, R.~{Ranasinghe},
  G.~{Dissanayake}, and T.~{Furukawa}, ``{Distance function based 6DOF
  localization for unmanned aerial vehicles in GPS denied environments},'' in
  \emph{IEEE International Conference on Intelligent Robots and Systems}, 2017,
  pp. 5292--5297.

\bibitem{Ryll2017ICRA}
M.~{Ryll}, G.~{Muscio}, F.~{Pierri}, E.~{Cataldi}, G.~{Antonelli},
  F.~{Caccavale}, and A.~{Franchi}, ``{6D physical interaction with a fully
  actuated aerial robot},'' in \emph{IEEE International Conference on Robotics
  and Automation}, 2017, pp. 5190--5195.

\bibitem{Lee2010CDC}
T.~{Lee}, M.~{Leok}, and N.~H. {McClamroch}, ``{Geometric tracking control of a
  quadrotor UAV on SE(3)},'' in \emph{49th IEEE Conference on Decision and
  Control}, 2010, pp. 5420--5425.

\bibitem{Faessler2018}
M.~{Faessler}, A.~{Franchi}, and D.~{Scaramuzza}, ``{Differential Flatness of
  Quadrotor Dynamics Subject to Rotor Drag for Accurate Tracking of High-Speed
  Trajectories},'' \emph{IEEE Robotics and Automation Letters}, vol.~3, no.~2,
  pp. 620--626, 2018.

\bibitem{Redmon2016CVPR}
J.~{Redmon}, S.~{Divvala}, R.~{Girshick}, and A.~{Farhadi}, ``{You Only Look
  Once: Unified, Real-Time Object Detection},'' in \emph{IEEE Conference on
  Computer Vision and Pattern Recognition}, 2016, pp. 779--788.

\bibitem{Jiao2019Access}
L.~{Jiao}, F.~{Zhang}, F.~{Liu}, S.~{Yang}, L.~{Li}, Z.~{Feng}, and R.~{Qu},
  ``{A Survey of Deep Learning-Based Object Detection},'' \emph{IEEE Access},
  vol.~7, pp. 128\,837--128\,868, 2019.

\bibitem{Girshick2015ICCV}
R.~{Girshick}, ``{Fast R-CNN},'' in \emph{IEEE International Conference on
  Computer Vision}, 2015, pp. 1440--1448.

\bibitem{Baca2020mrs}
\BIBentryALTinterwordspacing
T.~{Baca}, M.~{Petrlik}, M.~{Vrba}, V.~{Spurny}, R.~{Penicka}, D.~{Hert}, and
  M.~{Saska}, ``{The MRS UAV System: Pushing the Frontiers of Reproducible
  Research, Real-world Deployment, and Education with Autonomous Unmanned
  Aerial Vehicles},'' 2020. [Online]. Available:
  \url{https://arxiv.org/pdf/2008.08050}
\BIBentrySTDinterwordspacing

\bibitem{Pignaton2020ICUAS}
E.~{Pignaton de Freitas}, L.~A. L.~F. {da Costa}, C.~{Felipe Emygdio de Melo},
  M.~{Basso}, M.~{Rodrigues Vizzotto}, M.~{Schein Cavalheiro Corrêa}, and
  T.~{Dapper e Silva}, ``{Design, Implementation and Validation of a
  Multipurpose Localization Service for Cooperative Multi-UAV Systems},'' in
  \emph{2020 International Conference on Unmanned Aircraft Systems}, 2020, pp.
  295--302.

\bibitem{SanchezLopezJINT2017}
J.~L. {Sanchez-Lopez}, M.~{Molina}, H.~{Bavle}, C.~{Sampedro}, R.~A. {Suarez
  Fernandez}, and P.~{Campoy}, ``{A Multi-Layered Component-Based Approach for
  the Development of Aerial Robotic Systems: The Aerostack Framework},''
  \emph{Journal of Intelligent \& Robotic Systems}, vol.~88, no.~2, pp.
  683--709, 2017.

\bibitem{LimRAM2012}
H.~{Lim}, J.~{Park}, D.~{Lee}, and H.~J. {Kim}, ``{Build Your Own Quadrotor:
  Open-Source Projects on Unmanned Aerial Vehicles},'' \emph{IEEE Robotics
  Automation Magazine}, vol.~19, no.~3, pp. 33--45, 2012.

\bibitem{Furrer2016}
F.~Furrer, M.~Burri, M.~Achtelik, and R.~Siegwart, ``{RotorS -- A Modular
  Gazebo MAV Simulator Framework},'' in \emph{{Robot Operating System (ROS):
  The Complete Reference (Volume 1)}}, A.~{Koubaa}, Ed.\hskip 1em plus 0.5em
  minus 0.4em\relax Springer International Publishing, 2016, pp. 595--625.

\bibitem{Chaumette2006}
F.~{Chaumette} and S.~{Hutchinson}, ``{Visual servo control. I. Basic
  approaches},'' \emph{IEEE Robotics \& Automation Magazine}, vol.~13, no.~4,
  pp. 82--90, 2006.

\bibitem{Chaumette2007}
------, ``{Visual servo control. II. Advanced approaches [Tutorial]},''
  \emph{IEEE Robotics \& Automation Magazine}, vol.~14, no.~1, pp. 109--118,
  2007.

\bibitem{Siciliano2009}
B.~Siciliano, L.~Sciavicco, L.~Villani, and G.~Oriolo, \emph{{Robotics:
  Modelling, Planning and Control}}.\hskip 1em plus 0.5em minus 0.4em\relax
  Springer-Verlag, 2009, {ISBN:} 978-1846286414.

\bibitem{Lippiello2005}
V.~Lippiello, B.~Siciliano, and L.~Villani, ``{Eye-in-Hand/Eye-to-Hand
  Multi-Camera Visual Servoing},'' in \emph{Proceedings of the 44th IEEE
  Conference on Decision and Control}, 2005, pp. 5354--5359.

\bibitem{Silano2016}
G.~Silano and L.~Iannelli, ``{An educational simulation platform for GPS-denied
  Unmanned Aerial Vehicles aimed to the detection and tracking of moving
  objects},'' in \emph{IEEE Conference on Control Applications}, 2016, pp.
  1018--1023.

\bibitem{Bradski1998}
G.~R. Bradski, ``{Computer vision face tracking for use in a perceptual user
  interface},'' \emph{Intel Technology Journal}, 2nd Quarter 1998.

\bibitem{Rosten2006}
E.~Rosten and T.~Drummond, ``{Machine Learning for High-Speed Corner
  Detection},'' in \emph{9th European Conference on Computer Vision},
  L.~Ale{\v{s}}, B.~Horst, and A.~Pinz, Eds.\hskip 1em plus 0.5em minus
  0.4em\relax Springer Berlin Heidelberg, 2006, pp. 430--443.

\bibitem{Arefnezhad2018IJAT}
S.~{Arefnezhad}, A.~{Ghaffari}, A.~{Khodayari}, and S.~{Nosoudi}, ``{Modeling
  of Double Lane Change Maneuver of Vehicles},'' \emph{International Journal of
  Automotive Technology}, vol.~19, pp. 271--279, 2018.

\bibitem{IMUMathWorks}
\BIBentryALTinterwordspacing
{MathWorks Inc.}, ``{IMU Sensor Fusion with Simulink},'' 2020. [Online].
  Available:
  \url{https://www.mathworks.com/help/fusion/ug/imu-sensor-fusion-with-simulink.html}
\BIBentrySTDinterwordspacing

\bibitem{Bouabdallah2005AR}
S.~Bouabdallah, P.~Murrieri, and R.~Siegwart, ``{Towards Autonomous Indoor
  Micro VTOL},'' \emph{Autonomous Robots}, vol.~18, no.~2, pp. 171--183, 2005.

\bibitem{Corke2011}
P.~Corke, \emph{{Robotics, Vision and Control: Fundamental Algorithms in
  MATLAB}}.\hskip 1em plus 0.5em minus 0.4em\relax Springer, 2011, {ISBN
  978-3-319-54413-7}.

\bibitem{MatlabVRUserGuide}
\BIBentryALTinterwordspacing
{The MathWorks Inc.}, ``{Virtual Reality Toolbox: For use with MATLAB and
  Simulink},'' MathWorks official website. [Online]. Available:
  \url{http://cda.psych.uiuc.edu/matlab_pdf/vr.pdf}
\BIBentrySTDinterwordspacing

\bibitem{Silano_FramesAcquire2017}
\BIBentryALTinterwordspacing
G.~Silano, ``{Frames acquisition video},'' YouTube. [Online]. Available:
  \url{https://youtu.be/A70zed84zv0}
\BIBentrySTDinterwordspacing

\bibitem{DosAnjos2008}
A.~{dos Anjos} and H.~R. {Shahbazkia}, ``{BI-LEVEL IMAGE THRESHOLDING - A Fast
  Method},'' in \emph{Proceedings of the First International Conference on
  Bio-inspired Systems and Signal Processing}, vol.~2.\hskip 1em plus 0.5em
  minus 0.4em\relax SciTePress, 2008, pp. 70--76.

\bibitem{Viola2001}
P.~Viola and M.~Jones, ``{Rapid object detection using a boosted cascade of
  simple features},'' in \emph{Proceedings of the IEEE Computer Society
  Conference on Computer Vision and Pattern Recognition}, vol.~1, 2001, pp.
  511--518.

\bibitem{Xu2017}
Y.~{Xu}, G.~{Yu}, X.~{Wu}, Y.~{Wang}, and Y.~{Ma}, ``{An Enhanced Viola-Jones
  Vehicle Detection Method From Unmanned Aerial Vehicles Imagery},'' \emph{IEEE
  Transactions on Intelligent Transportation Systems}, vol.~18, no.~7, pp.
  1845--1856, 2017.

\bibitem{Gonzalez2019}
P.~Martinez-Gonzalez, S.~Oprea, A.~Garcia-Garcia, A.~Jover-Alvarez,
  S.~Orts-Escolano, and J.~Garcia-Rodriguez, ``{UnrealROX: an extremely
  photorealistic virtual reality environment for robotics simulations and
  synthetic data generation},'' \emph{Virtual Reality}, 2019.

\bibitem{Artner2008}
N.~M. Artner and W.~Burger, ``{A Comparison of Mean Shift Tracking Methods},''
  in \emph{2017 International Conference on Unmanned Aircraft Systems}, 2008,
  pp. 197--204.

\bibitem{Bouabdallah2005ICRA}
S.~{Bouabdallah} and R.~{Siegwart}, ``{Backstepping and Sliding-mode Techniques
  Applied to an Indoor Micro Quadrotor},'' in \emph{Proceedings of the 2005
  IEEE International Conference on Robotics and Automation}, 2005, pp.
  2247--2252.

\bibitem{Stevens}
B.~L. Stevens, F.~L. Lewis, and E.~N. Johnson, \emph{{Aircraft control and
  simulation: dynamics, controls design, and autonomous systems}}.\hskip 1em
  plus 0.5em minus 0.4em\relax John Wiley \& Sons, 2015, {ISBN:
  978-1-118-87098-3}.

\bibitem{Nascimento2019}
T.~P. Nascimento and M.~Saska, ``{Position and attitude control of multi-rotor
  aerial vehicles: A survey},'' \emph{Annual Reviews in Control}, vol.~48, pp.
  129--146, 2019.

\bibitem{Pestana2014}
J.~Pestana, J.~L. Sanchez-Lopez, S.~Saripalli, and P.~Campoy, ``{Computer
  vision based general object following for {GPS}-denied multirotor unmanned
  vehicles},'' in \emph{IEEE American Control Conference}, 2014, pp.
  1886--1891.

\bibitem{Dief2015}
T.~N. Dief and S.~Yoshida, ``{Review: Modeling and Classical Controller Of
  Quad-rotor},'' \emph{International Journal of Computer Science and
  Information Technology \& Security}, vol.~5, no.~4, pp. 314--319, 2015.

\bibitem{Silano2017_referenceGenerator}
\BIBentryALTinterwordspacing
G.~Silano, ``{Vision-based target tracking scenario. The drone dynamics and the
  control system are neglected to evaluate the performance of the reference
  generator},'' YouTube. [Online]. Available:
  \url{https://youtu.be/qAtndBIwdas}
\BIBentrySTDinterwordspacing

\bibitem{Silano2017_droneFollowsACar}
\BIBentryALTinterwordspacing
------, ``{Vision-based target tracking scenario},'' YouTube. [Online].
  Available: \url{https://youtu.be/b8mTHRkRDmA}
\BIBentrySTDinterwordspacing

\bibitem{Silano2017_carPartiallyCover}
\BIBentryALTinterwordspacing
------, ``{Vision-based target tracking scenario when the target is partially
  covered},'' YouTube. [Online]. Available: \url{https://youtu.be/RjXBtPqZZBc}
\BIBentrySTDinterwordspacing

\bibitem{worldChanged}
\BIBentryALTinterwordspacing
------, ``Quad-rotor following a car that moves along a nontrivial path in the
  vr\_octavia scenario,'' YouTube. [Online]. Available:
  \url{https://youtu.be/m43Zadq-6XM}
\BIBentrySTDinterwordspacing

\end{thebibliography}



\begin{IEEEbiography}[{\includegraphics[width=1in,height=1.25in,clip,keepaspectratio] {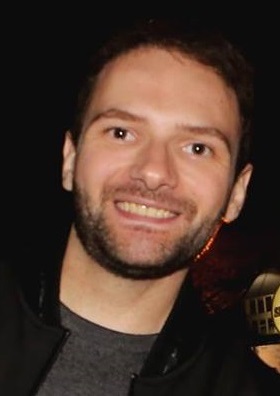}}]{Giuseppe Silano} (S'16) received the bachelor and master degrees in computer (2012) and electronic engineering (2016), respectively, and the PhD degree in information engineering (2020) from the University of Sannio, Italy. From June 2020, he is with the Czech Technical University in Prague where he holds a post-doctoral research fellowship position. From March to November 2019, he was a visiting student at LAAS--CNRS, in Toulouse, France. His research interests are in simulation and control, temporal logic, model predictive control, software-in-the-loop, and planning for micro aerial vehicles. He was among the finalists of the ``Aerial robotics control and perception challenge'', the Industrial Challenge of the 26th Mediterranean Conference on Control and Automation (MED'18), and among the participants of the LAAS Team selected as a finalist of the Mohammed Bin Zayed Robotics Competition (MBZIRC) 2020. He is a member of the IEEE Control System Society (CSS) and IEEE Robotics and Automation Society (RAS).
\end{IEEEbiography}
\begin{IEEEbiography}[{\includegraphics[width=1in,height=1.25in,clip,keepaspectratio] 
{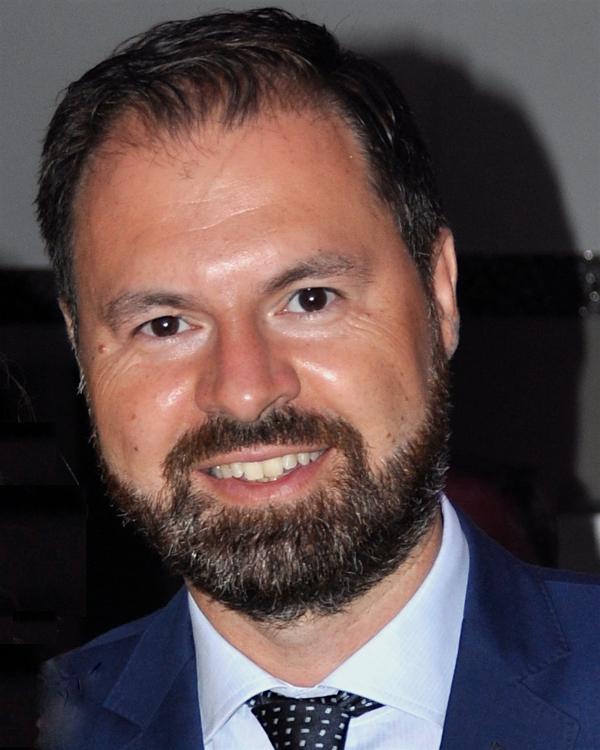}}]{Luigi Iannelli} (S'00-M'02-SM'12) received the master's degree (Laurea) in computer engineering from the University of Sannio, Italy, in 1999 and the PhD degree in information engineering from the University of Napoli Federico II, Italy, in 2003. After a postdoc position with the University of Napoli Federico II, he joined the University of Sannio as an assistant professor, and since 2016 he has been an associate professor of automatic control. He held visiting researcher positions in the Royal Institute of Technology, Sweden, and at the Johann  Bernoulli Institute of Mathematics and Computer Science, University of Groningen, The Netherlands. His current research interests include analysis and control of switched and nonsmooth systems, stability analysis of piecewise-linear systems, smart grid control, and applications of control theory to power electronics and UAVs. He was co-editor of the book Dynamics and Control of Switched Electronic Systems (Springer, 2012). He is a senior member of the IEEE. 
\end{IEEEbiography}

\EOD

\end{document}